\documentclass{article}
\usepackage{spconf}
\usepackage[latin9]{inputenc}
\usepackage[english]{babel}
\usepackage{array}
\usepackage{amsmath}
\usepackage{amssymb}
\usepackage{graphicx}
\usepackage{nomencl}
\usepackage{tablefootnote}
\usepackage{subcaption}
\usepackage{placeins}
\usepackage{makecell}
\usepackage{color}

\providecommand{\makenomenclature}{\makeglossary}
\makenomenclature
\usepackage[unicode=true]{hyperref}
\usepackage{breakurl}

\makeatletter

\providecommand{\tabularnewline}{\\}

\usepackage{babel}
\AtBeginDocument{

}

\makeatother
 
\begin{document}

\title{Microscopy Cell Segmentation via Convolutional LSTM Networks}

\name{Assaf Arbelle and Tammy Riklin Raviv\sthanks{This study was partially supported by the
    Negev scholarship at Ben-Gurion University (A.A.); The Kreitman
    School of Advanced Graduate Studies (A.A) ; The Israel Ministry of Science, Technology and Space
 (MOST 63551 T.R.R.)}}

\address{The Department of Electrical and Computer Engineering \\ The Zlotowski Center for Neuroscience\\ Ben-Gurion University of the Negev \\ } 
\maketitle
\begin{abstract}
Live cell microscopy sequences exhibit complex spatial structures and complicated temporal behaviour, making  their analysis a challenging task. Considering cell segmentation problem, which plays a significant role in the analysis, the spatial properties of the data can be captured using Convolutional Neural Networks (CNNs). Recent approaches show promising segmentation results using convolutional encoder-decoders such as the U-Net. Nevertheless,  these methods are limited by their inability to incorporate temporal information, that can facilitate segmentation of individual touching cells or of cells that are partially visible. In order to exploit cell dynamics we propose a novel segmentation architecture which integrates Convolutional Long Short Term Memory (C-LSTM) with the U-Net. The network's unique architecture allows it to capture multi-scale, compact, spatio-temporal encoding in the C-LSTMs memory units. The method was evaluated on the Cell Tracking Challenge and achieved state-of-the-art results (1st on Fluo-N2DH-SIM+ and 2nd on DIC-C2DL-HeLa datasets)
The code is freely available at: \url{https://github.com/arbellea/LSTM-UNet.git}

% To capture cell dynamics and at the same time allow multi-scale spatial encoding of the data we propose a novel segmentation approach which is based on Convolutional Long Sort Term Memory (C-LSTM) network integration both the spatial and temporal aspects of the data. The network's unique architecture allows it to capture multi-scale, compact, spatio-temporal encoding in the C-LSTMs memory units. Promising results, surpassing the state-of-the-art, are presented.  
%The code is freely available at: ANONYMOUS
\end{abstract}

\section{Introduction \label{sec:Introduction}}
Live cell microscopy imaging is a powerful tool and an important part of the biological research process. % Nevertheless, the research is hindered by the exhaustive process of annotating the acquired sequences. 
The automatic annotation of the image sequences is crucial for the quantitative analysis of properties such as cell size, mobility, and protein levels. 
Recent image analysis approaches have shown the
strengths of Convolutional Neural Networks (CNNs) which surpass state-of-the-art methods in virtually all fields, such as object classification \cite{krizhevsky12}, detection \cite{redmon2016}, semantic segmentation~\cite{long2015}, and many other tasks. Attempts at cell segmentation using CNNs include \cite{arbelle2017SAN,kraus2016,Ronneberger15}. All these methods, however, are trained on independent, non sequential, frames and do not incorporate any temporal information which can potentially facilitate segmentation in cases of neighboring cells that are hard to separate or when a cell partially vanishes. The use of temporal information by combining tracking information from individual cells to support segmentation decisions has been shown to improve results for non deep learning methods \cite{Amat14,Arbelle15,Schiegg14,arbelle2018probabilistic} but have not yet been extensively examined in a deep learning approaches. 
%There are, however, non-deep learning methods that utilize the temporal information by combining tracking information from individual cells to support segmentation decisions \cite{Amat14,Arbelle15,Schiegg14,arbelle2018probabilistic}.

A Recurrent Neural Network (RNN) is an artificial neural network equipped with feed-back connections. This unique architecture makes it suitable for the analysis of dynamic behavior. A special variant of RNNs is Long Short Term Memory (LSTM), which includes an internal memory state vector with gating operations thus stabilizing the training process \cite{hochreiter1997LSTM}. Common LSTM based applications include natural language processing (NLP) \cite{vinyals2015grammar},  audio processing \cite{sak2014long} and image captioning \cite{xu2015show}.

Convolutional LSTMs (C-LSTMs) accommodate locally spatial information in image sequences by replacing matrix multiplication with convolutions  \cite{xingjian15convLSTM}.  The C-LSTM has recently been used to address the analysis of both temporal image sequences, such as next frame prediction \cite{lotter2016deep}, and  volumetric data sets  \cite{chen2016combining,stollenga2015parallel}. In \cite{stollenga2015parallel} C-LSTM is applied in multiple directions for the segmentation of 3D data represented as a stack of 2D slices.
Another approach for 3D brain structure segmentation is proposed in \cite{chen2016combining}, where each slice is separately fed into a U-Net architecture, and only the output then fed into bi-directional C-LSTMs.

In this paper we introduce the integration of C-LSTMs into an encoder-decoder structure (U-Net) allowing compact spatio-temporal representations in multiple scales. 
We note that, unlike \cite{chen2016combining} which was designed and evaluated on 3D brain segmentation, the proposed novel architecture is an intertwined composition of the two concepts rather than a pipeline. Furthermore, since our method is designed for image sequence segmentation which can be very long the bi-directional C-LSTM is not computationally feasible.  Our framework is assessed using time-lapse microscopy data where both cells' dynamics and their spatial properties should be considered. Specifically, we tested our method on the Cell Tracking Challenge: \url{http://www.celltrackingchallenge.net}. Our method was ranked in the top three by the challenge organizers on the several submitted data sets, specifically on the fluorescent  simulated dataset (Fluo-N2DH-SIM+) and the  differential interference contrast (DIC-C2DL-HeLa)  sequences which are difficult to segment. 

The rest of the paper is organized as follows. Section~\ref{sec:Proposed-Algorithm} presents a probabilistic formulation of the problem and elaborates on the proposed network. Technical aspects are detailed in Section~\ref{sec:Implementation Details}. In Section~\ref{sec:Experiments} we demonstrate the strength of our method, presenting state-of-the-art cell segmentation results. We conclude in Section~\ref{sec:Summary}.

%There are several tools available for automatic and semi-automatic segmentation such as Ilastik \citep{Sommer2011} and CellProfiler \citep{Carpenter06}. However these tools require tuning of multiple parameters, and may produce sub-optimal results. 

\section{Methods}\label{sec:Proposed-Algorithm}
\subsection{Network Architecture}\label{subsec:Network-Architecture} 
\begin{figure}
\noindent \begin{centering}
\includegraphics[trim={8bp 595bp 435bp 10bp},clip, width=0.95\columnwidth]{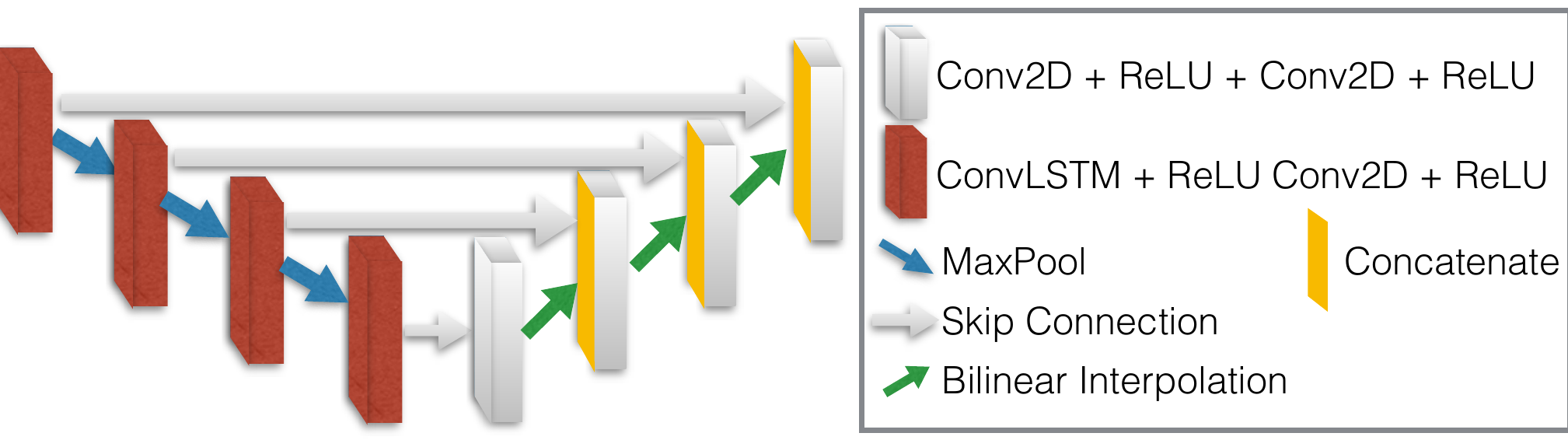}  %trim={30bp 140bp 168bp 60bp},clip, %trim={left, bottom, right, top}
\par\end{centering}
\protect\caption{The U-LSTM network architecture. The down-sampling path (left) consists of a C-LSTM layer followed by a convolutional layer with ReLU activation, the output is then down-sampled using max pooling and passed to the next layer. The up-sampling path (right) consists of a concatenation of the input from the lower layer with the parallel layer from the down-sampling path followed by two convolutional layers with ReLU activations.  \label{fig:Architecture}}
\end{figure}

The proposed network %$f_\Theta$ 
incorporates C-LSTM \cite{xingjian15convLSTM} blocks into the U-Net \cite{Ronneberger15} architecture. This combination, as suggested here, is shown to be powerful. The U-Net architecture, built as an encoder-decoder with skip connections, enables to extract meaningful descriptors at multiple image scales. However, this alone does not account for the cell specific dynamics that can significantly support the segmentation. %which siginifcantly supports the segmentation.
The introduction of C-LSTM blocks into the network allows considering past cell appearances at multiple scales by holding their compact representations in the C-LSTM memory units.
We propose here the incorporation of C-LSTM layers in every scale of the encoder section of the U-Net. Applying the CLSTM on multiple scales is essential for cell microscopy sequences (as opposed to brain slices as in \cite{chen2016combining}) since the frame to frame differences might be at different scales, depending on cells' dynamics. Moreover, in contrast to brain volume segmentation \cite{chen2016combining} the microscopy sequence can be of arbitrary length, making the use of bi-directional LSTMs computationally impractical and the cells can move at different speeds and the changes are not normally smooth. 
The comparison to other alternatives is presented in Section~\ref{ArchitectureSelection}. 
The network is fully convolutional and, therefore, can be used with any image size\footnote{In order to avoid artefacts it is preferable to use image sizes which are multiples of eight due to the three max-pooling layers.} during both training and testing. Figure~\ref{fig:Architecture} illustrates the network architecture detailed in Section~\ref{sec:Implementation Details}.
 \subsection{Formulation}\label{subsec:Formulation} 
We address individual cells' segmentation from microscopy sequences. The main challenge in this type of problems is not only foreground-background classification but also the separation of adjacent cells. We adopt the  weighted distance loss as suggested by \cite{Ronneberger15}. The loss is designed to enhance individual cells' delineation by a partitioning of the $d$ dimensional (2 or 3) image domain $\Omega \in \mathbb{R}^{d}$ into two classes: foreground and background, such that pixels which are near the boundaries of two adjacent cells are given higher importance.
We set $\mathcal{C} = \{0,1\}$  to denote these classes, respectively. Let $\{I_t\}_{t=1} ^{T}$ be the input image sequence of length $T$, % with size $M\times N$, 
where $I_t: \Omega\rightarrow \mathbb{R} $ is a grayscale image. 
The network is composed of two sections of $L$ blocks each, the encoder recurrent block $E_{\theta_l}^{\{l\}}(\cdot)$ and the decoder block  $D_{\theta_l}^{\{l\}}(\cdot)$ where $\theta_l$ are the network's parameters. 
The input to the C-LSTM encoder layer $l\in [0,\ldots,L-1]$ at time $t\in T$ includes the down-sampled output of the previous layer, the output of the current layer at the previous time-step and the C-LSTM memory cell. We denote these three inputs as $x_t^{\{l\}}$, $h_{t-1}^{\{l\}}$, $c_{t-1}^{\{l\}}$ respectively. Formally we define: 
\begin{equation}
(h_{t}^{\{l\}}, c_{t}^{\{l\}})=E_{\theta_l}^{\{l\}}(x_t^{\{l\}}, h_{t-1}^{\{l\}}, c_{t-1}^{\{l\}})
\end{equation}
where,
\begin{equation}
    x_t^{\{l\}}= 
\begin{cases}
    I_t,&  l=0\\
    MaxPool(h_t^{\{l-1\}}), & 0<l<L

\end{cases}
\end{equation}

The inputs to the decoder layers $l\in [L,\ldots,2L-1]$  are the up-sampled \footnote{We use bi-linear interpolation} output of the previous layer and the output of the corresponding layer from the encoder denoted by $y_t^{\{l\}}$ and  $h_t^{\{2L-1-l\}}$ respectively. We denote the decoder output as $z_{t}^{\{l\}}$. Formally,

\begin{equation}
    y_t^{\{l\}}= 
\begin{cases}
    h_t^{\{l-1\}},&  l=L\\
    UpSample(z_t^{\{l-1\}}), & L<l<2L-1

\end{cases}
\end{equation}
\begin{equation}
z_t^{\{l\}} = D_{\theta_l}(y_t^{\{l\}}, h_t^{\{2L-1-l\}})
\end{equation}

We define a network  $f_{\Theta}$ with parameters $\Theta$ as  the composition of $L$ encoder blocks followed by $L$ decoder blocks, and denote $\Theta := \{\theta_l\}_{l=0}^{2L-1}$ . Note that the encoder blocks, $E_{\theta_l}^{\{l\}}$,  encode high-level \textbf{spatio-temporal} features at multiple scales and the decoder blocks, $D_{\theta_l}^{\{l\}}$, refines that information into a full scale segmentation map.

\begin{equation}
o_t \overset{\Delta}{=} f_{\Theta}= z_t^{\{2L-1\}}
\end{equation}
We set the final output as a $|\mathcal{C}|$-dimensional feature vector corresponding to each input pixel $\mathbf{v}\in \Omega$. 
We define the segmentation as the pixel label probabilities using the softmax equation: \begin{equation}
p(c|o_t(\mathbf{v}))=\frac{\exp\{[o_t(\mathbf{v})]_c\}}{\sum_{c'\in\mathcal{C}}\exp\{[o_t(\mathbf{v})]_{c'}\}}, ~~~ c\in\mathcal{C}\label{softmax}
\end{equation}
The final segmentation is defined as follows:
\begin{equation}
\hat{\Gamma}_t = \arg_{c\in C} \max p(c|o_t(\mathbf{v}))
\end{equation}
Each connected component of the foreground class is given a unique label and is considered an individual cell. 

\subsection{Training and Loss}\label{subsec:Training}

During the training phase the network is presented with a full sequence and manual annotations $\{I_t,\Gamma_t\}_{t=1}^T$, where $\Gamma_t: \Omega\rightarrow[0,1]$ are the ground truth (GT) labels. 
%As in the U-Net \cite{Ronneberger15}, given $\Gamma_t$, in order to ensure single cell separation, we define a weight map $w_d$ that accounts for the Euclidean distances of the image pixels with respect to the nearest and the second nearest cells' pixel, $d_1$ and $d_2$ respectively.
%\begin{equation}
%\label{eg:DistanceWeight}
%w_d = w_0 \exp\left\lbrace-({d_1+d_2)^2}/{\sigma_0}^2\right\rbrace
%\end{equation}
The network is trained using Truncated Back Propagation Through Time (TBPTT) \cite{williams1990efficient}. At each back propagation step the network is unrolled to $\tau$ time-steps. The loss is defined using the distance weighted cross-entropy loss as proposed in the original U-Net paper \cite{Ronneberger15}. The loss imposes separation of cells by introducing an exponential penalty factor wich is  proportional to the distance of a pixel from its nearest and second nearest cells' pixels. Consequently, pixels which are located between two adjacent cells are given significant importance whereas pixels further away from the cells have a minor effect on the loss. A detailed discussion on the weighted loss can be found in the original U-Net paper \cite{Ronneberger15}

\section{Implementation Details}\label{sec:Implementation Details}

\subsection{Architecture}
The network comprises $L=4$ encoder and decoder blocks. Each block in the encoder section is composed of C-LSTM layer, leaky ReLU, convolutional layer, batch normalization \cite{ioffe2015batchnorm}, leaky ReLU and finally down-sampled using maxpool operation. The decoder blocks consist of a bi-linear interpolation, a concatenation with the parallel encoder block and  an followed by two convolutional layer, batch normalization \cite{ioffe2015batchnorm}, and leaky ReLU.  All convolutional layers use kernel size $3\times 3$ with layer depths $(128,256,512,1024)$. All maxpool layers use kernel size $2\times 2$ without overlap. All C-LSTM kernels are of size $3\times 3$ and $5
\times 5$ respectively with layer depths $(1024,512,256,128)$. The last convolutional layer uses kernel size $1\times 1$ with depth $2$ followed by a softmax layer to produce the final probabilities (see Figure~\ref{fig:Architecture}).

\subsection{Training Regime}

We trained the networks for approximately $100K$ iterations with an RMS-Prop optimizer \cite{HintonRMSPROP} with learning rate of 0.0001. The unroll length parameter was set to $\tau =5$ was set (Section~\ref{subsec:Training}) and the batch size was set to three sequences. %The parameter for the weighted loss function were set to: $w_c=1$, $w_0=10$ and $\sigma_0=5$ for all experiments. 

\subsection{Data}
The images were annotated using two labels for the background and cell nucleus. In order to increase the variability, the data was randomly augmented spatially and temporally by: 1) random horizontal and vertical flip, 2) random $90^o$ rotation 3) random crop of size $160\times 160$ 4) random sequence reverse ($[T, T-1,\ldots,2,1]$), 5) random temporal down-sampling by a factor of $k \in [0,4]$, 6) random affine and elastic transformations.% The total training set following the proposed augmentation is hundreds of times larger than the original set. 
We note that the gray-scale values are not augmented as they are biologically meaningful.

%These annotations are presented visualization as RGB images, red indicating the background, green indicating the cell nucleus and blue indicating the contour.

\section{Experiments and Results}\label{sec:Experiments}
\begin{table}
\begin{small}
\begin{centering}
\begin{tabular}{c|>{\centering}p{0.16\columnwidth}|>{\centering}p{0.16\columnwidth}|>{\centering}p{0.16\columnwidth}}
\hline 
\textbf{Architecture} & \small{EncLSTM} & \small{DecLSTM} & \small{FullLSTM}
\tabularnewline
\hline
\textbf{SEG} & 0.874 & 0.729 &0.798 
\tabularnewline
~&($\pm$0.011)&($\pm$0.166)&($\pm$0.094)
\tabularnewline
\hline
\end{tabular}
\par\end{centering}
\caption{\label{tab:ArchitectureExp}\textbf{Architecture Experiments} Comparison of three variants of the proposed network incorporating C-LSTMs in: (1). the encoder seuction (EncLSTM) (2). the decoder section (DecLSTM) (3). both the encoder and decoder sections (FullLSTM).  The training procedure was repeated three times, mean and standard deviation are presented}
\end{small}
\end{table}
 \begin{table*}
\begin{small}
\begin{centering}
\begin{tabular}{|c||>{\centering}p{0.09\paperwidth}|>{\centering}p{0.09\paperwidth}||>{\centering}p{0.09\paperwidth}|>{\centering}p{0.09\paperwidth}|>{\centering}p{0.09\paperwidth}|}
\hline 
\textbf{Dataset} & EncLSTM (BGU-IL$^{(4)}$)& DecLSTM (BGU-IL$^{(2)}$)  & \textbf{First} & \textbf{Second} & \textbf{Third} 
\tabularnewline
\hline
\hline

\textbf{Fluo-N2DH-SIM+} &\textbf{0.811 (1st)}& \textbf{0.802 (3rd)} & \textbf{0.811}$^*$ & 0.807$^{(b)}$ & \textbf{0.802}$^{**}$ \tabularnewline

\hline

\textbf{DIC-C2DL-HeLa}&\textbf{0.793 (2nd)} & 0.511 (5th) &0.814$^{(a)}$ & \textbf{0.793}$^*$ & 0.792$^{(b)}$ \tabularnewline

\hline

\textbf{PhC-C2DH-U373}& 0.842 (5th) & -- & 0.924$^{(c)}$ &0.922$^{(b)}$ & 0.920$^{(d)}$ \tabularnewline

\hline

\textbf{Fluo-N2DH-GOWT1}& 0.850 (8th) & 0.854 (7th) & 0.927$^{(f)}$ &0.894$^{(b)}$ & 0.893$^{(g)}$ \tabularnewline

\hline
\textbf{Fluo-N2DL-HeLa}& 0.811 (8th) &0.839 (6th) & 0.903$^{(b)}$ &0.902$^{(d)}$ & 0.900$^{(e)}$ \tabularnewline

\hline

\end{tabular}
\par\end{centering}
\caption{\label{tab:Results}\textbf{Quantitative Results:} Method evaluation on the submitted dataset (challenge set) as evaluated and published by the Cell Tracking Challenge organizers \cite{Ulman17}.  Our methods EncLSTM and DecLSTM are referred to here as BGU-IL$^{(4)}$ and BGU-IL$^(2)$ respectively.
Our method ranked  first on the Fluo-N2DH-SIM+ and second on the DIC-C2DL-HeLa dataset.
The three columns on the right report the results of the top three methods as named by the challenge organizers. The measure is explained in Section~\ref{subsec:Evaluation}. The superscript (a-g) represent different methods: 
(*). EncLSTM \textbf{ours}, (**) DecLSTM \textbf{ours}, (a).TUG-AT, (b). CVUT-CZ, (c). FR-Fa-GE, (d). FR-Ro-GE (Original UNet \cite{Ronneberger15}), (e). KTH-SE, (f) LEID-NL.}
\end{small}
%\vspace{-0.2cm}
\end{table*}

\subsection{Evaluation Method}\label{subsec:Evaluation}
The method was evaluated using the scheme proposed
in the online version of the Cell Tracking Challenge. Specifically, SEG for segmentation \cite{Ulman17}. % and  TRA for tracking, as defined in \cite{Ulman17}. 
%The tracking was done by associating cells with overlap greater than half in consecutive frames. We chose the most naive tracking method, to show that our results are based only on segmentation accuracy. 
The SEG measure is defined as the mean Jaccard index
$\frac{|A\cap B|}{|A\cup B|}$ of a pair of ground truth label $A$ and its corresponding segmentation $B$. A segmentation is considered a match if $|A \cap B|> \frac{1}{2}|A|$ . 

 %The TRA uses the  Acyclic Oriented Graph Matching ($AOGM$) which assesses how difficult it is to transform a computed graph into a given ground-truth graph. The TRA measure is defined as follows: $$TRA = 1- \min{\{AOGM, AOGM_0\}}/AOGM_0\label{eg:TRA}$$ where $AOGM_0$ is the $AOGM$ value required for creating the reference graph from scratch.  OP is defined as the mean of TRA and SEG.

\subsection{Architecture Selection:}
 \label{ArchitectureSelection}
 We propose integrating the CLSTM into the U-Net by substituting the convolutional layers of the encoder section with C-LSTM layers (referred to as EncLSTM). 
In this section we compare this architecture with two alternatives by substituting: 1) the convolutional layers of the decoder section (referred to as DecLSTM); 2) the convolutional layers of both the decoder and encoder sections (referred to as FullLSTM). All three networks were trained simultaneously with identical inputs. Due to the limited size of the training set, the networks were trained on the Fluo-N2DH-SIM+ datasets and tested on the similar Fluo-N2DH-GOWT1 datasets from the training set of the Cell Tracking Challenge. The results as presented  Table~\ref{tab:ArchitectureExp} show an advantage for the proposed architecture. Howerver, the dominance of the EncLSTM with respect to the DecLSTM is not conclusive as is demonstrated by the result for the cell tracking challenge discussed next. We further note that a comparison to the original U-Net, without LSTM, is obtained in the challenge results and is referred to by the challenge organizers as FR-Ro-Ge. The method is labelled in Table~\ref{tab:Results} with the superscript $(d)$.

\subsection{Cell Tracking Challenge Results:}
 \label{ChallengeExperiments}

\begin{figure}
\begin{centering}
\setlength{\tabcolsep}{0mm}
\renewcommand{\arraystretch}{0.2}

\begin{tabular}{c}

\begin{subfigure}[b]{.03\linewidth}
\begin{tabular}{cc}
\rotatebox{90}{~~~~~~~~~Full Image}\\
\rotatebox{90}{~~~~Zoom}\\
~\\~
\end{tabular}
\end{subfigure}
\begin{subfigure}[b]{.31\linewidth}
\begin{tabular}{cc}
\includegraphics[trim={0bp 95bp 0bp 95bp},clip, width=1\columnwidth]{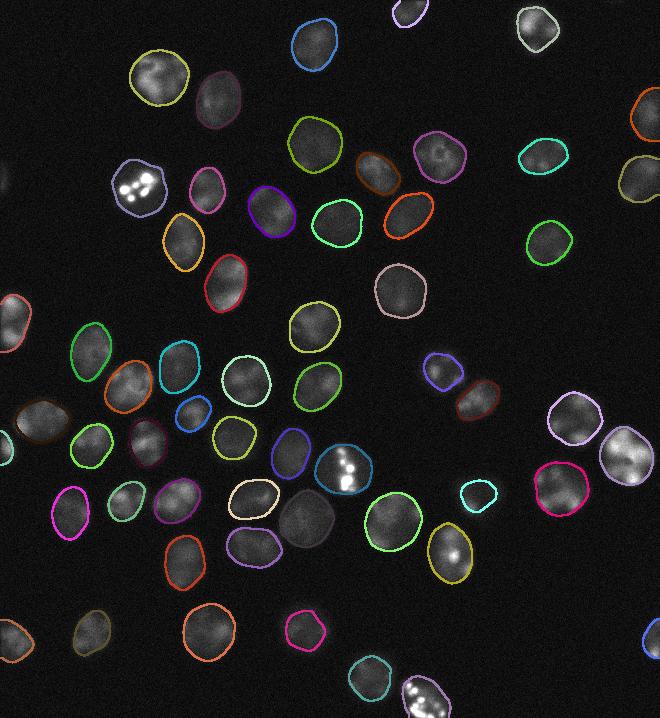}   \\
\includegraphics[trim={100bp 484bp 435bp 134bp},clip,width=1\columnwidth]{./Images/SIM-01RES_frame_91}  \\~\\
Fluo-N2DH-SIM+
%718x660 --> 528x660
\end{tabular}
\end{subfigure}
\begin{subfigure}[b]{.31\linewidth}
\begin{tabular}{c}
\includegraphics[trim={0bp 0bp 0bp 102bp},clip, width=1\columnwidth]{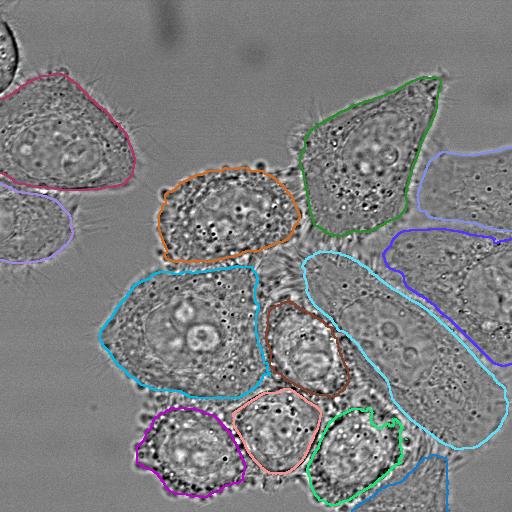}     \\
\includegraphics[trim={80bp 102bp 109bp 152bp},clip,width=1\columnwidth]{./Images/DIC-01RES_frame_13}\\~\\
DIC-C2DH-HeLa
%576x720 
\end{tabular}
\end{subfigure}
\begin{subfigure}[b]{.31\linewidth}
\begin{tabular}{c}
\includegraphics[trim={19bp 0bp 18bp 0bp},clip, width=1\columnwidth]{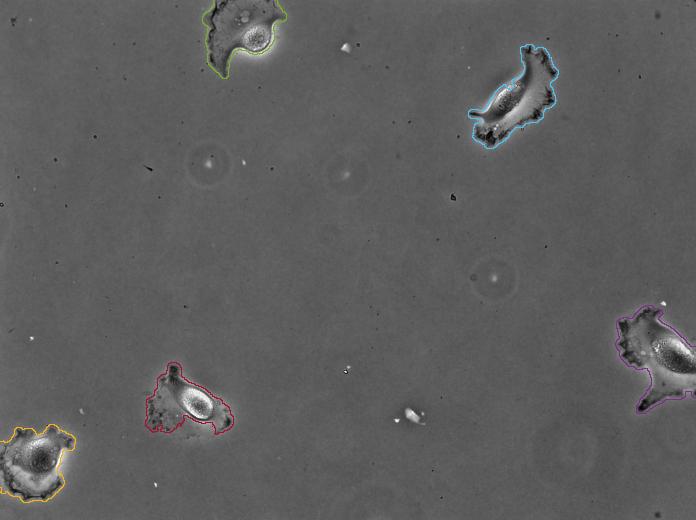}   \\
\includegraphics[trim={460bp 385bp 111bp 35bp},clip,width=1\columnwidth]{./Images/U373t108}   \\~\\
PhC-C2DH-U373x
%520x696 >>520x659
\end{tabular}
\end{subfigure}
\tabularnewline
\end{tabular}
\par\end{centering}
\caption{\label{fig:Results}
Examples of the segmentation results for three of the data sets
The top and bottom row show the full image and  a zoom-in of a specific area respectively.  The columns from left to right show the Fluo-N2DH-SIM+, DIC-C2DH-HeLa and PhC C2DH-U373 datasets. Note that 1. even though the cells are touching, they are correctly separated; 2. The complex shape and texture of the cell is correctly segmented.}

%\vspace{-0.4cm}
\end{figure}
%\vspace{-0.3cm}

Two variants of the method, EncLSTM and DecLSTM, were applied to five data sets: Fluo-N2DH-SIM+, DIC-C2DL-HeLa, PhC-C2DH-U373, Fluo-N2DH-GOWT1, Fluo-N2DL-HeLa. The results were submitted to the Cell Tracking Challenge. The proposed Enc-LSTM and Dec-LSTM were ranked 1st and 3rd, respectively, out of 20, for the Fluo-N2DH-SIM+ data set and 2nd out of 10 (EncLSTM) for the DIC-C2DL-HeLa dataset. We note that for the other three datasets - training data were significantly smaller and this might explain the inferior results we received. A possible solution to this problem is using adversarial loss as suggested in \cite{arbelle2017SAN}.
In general, 20 different methods have been submitted to the challenge, including the original U-Net (FR-Ro-GE) \cite{Ronneberger15} and TUG-AT \cite{Payer2018}. 
 The latter  independently and simultaneously proposed to utilize the U-Net architecture while introducing C-LSTM layers on the skip connections.   Table~\ref{tab:Results} reports our results in comparison to the three leading methods (including ours) provided by the challenge organizers. Visualizations of the results are presented in Fig~\ref{fig:Results} and in  \href{https://youtu.be/IHULAZBmoIM}{https://youtu.be/IHULAZBmoIM}. The quantitative results for top three leading methods are also publicly available at the \href{http://www.celltrackingchallenge.net/latest-ctb-results/}{Cell Tracking Challenge web site}..

\vspace{-0.175cm}
\section{Summary}\label{sec:Summary}

\vspace{-0.1cm}
Time-lapse microscopy cell segmentation is, inherently, a spatio-temporal task. Human annotators frequently rely on temporal queues in order to accurately separate neighbouring cells and detect partially visible cells. In this work, we demonstrate the strength of integrating temporal analysis, in the form of C-LSTMs, into a well established network architecture (U-Net) and examined several alternative combinations. The resulting novel architecture is able to extract meaningful features at multiple scales and propagate them through time. This enables the network to accurately segment cells in difficult scenarios where the temporal queues are crucial. Quantitative analysis shows that our method achieves state-of-the-art results (Table~\ref{tab:Results}) ranking 1st and 2nd place in the Cell Tracking Challenge\footnote{Based on the October 15th, 2018 ranking.}. Moreover, the results reported in Table~\ref{tab:ArchitectureExp}
demonstrate the proposed network ability 
to generalize from simulated training data 
to real data. This may imply that one can reduce and even eliminate the need for extensive manually annotated data. We further plan on incorporating adversarial loss to weaken the dependancy on training set size as in \cite{arbelle2017SAN}. 

%The code, implemented in Tensorflow, is freely available at: Anonymous.

%\subsubsection*{Acknowledgements}
%This study was partially supported by the Negev scholarship at Ben-Gurion University)(A.A); The Kreitman School of Advanced Graduate Studies (A.A); Israel Science Foundation (1638/16)(T.RR); The Israel Defense Forces  (IDF) Medical Corps and Directorate of Defense Research \& Development, Israeli Ministry of Defense (IMOD DDR\&D) (T.RR); The Israeli ministry of science, technology and space (63551) (T.RR)

%\bibliographystyle{abbrv}

\FloatBarrier

\bibliographystyle{IEEEbib}
\bibliography{DeepSeg}

\end{document}